\documentclass[english,preprint, 3p, twocolumn, number, sort]{elsarticle}

\usepackage[T1]{fontenc}
\usepackage[latin9]{inputenc}
\usepackage{float}
\usepackage{amstext}
\usepackage{soul}
\usepackage{etoolbox}

\usepackage{array}
\newcolumntype{L}[1]{>{\raggedright\let\newline\\\arraybackslash\hspace{0pt}}m{#1}}
\newcolumntype{C}[1]{>{\centering\let\newline\\\arraybackslash\hspace{0pt}}m{#1}}
\newcolumntype{R}[1]{>{\raggedleft\let\newline\\\arraybackslash\hspace{0pt}}m{#1}}

\makeatletter

\newcommand{\lyxmathsym}[1]{\ifmmode\begingroup\def\b@ld{bold}
  \text{\ifx\math@version\b@ld\bfseries\fi#1}\endgroup\else#1\fi}

\def\ps@pprintTitle{%
 \let\@oddhead\@empty
 \let\@evenhead\@empty
 \def\@oddfoot{}%
 \let\@evenfoot\@oddfoot}

\makeatother

\usepackage{babel}
\usepackage{amssymb,amsmath,amsthm}
\usepackage{graphicx,epstopdf,color}
\usepackage{algorithm}
\usepackage{algpseudocode}
\usepackage{caption}
\usepackage{subcaption}
\usepackage{wrapfig}
\usepackage[dvipsnames]{xcolor}

\algdef{SE}[DOWHILE]{Do}{doWhile}{\algorithmicdo}[1]{\algorithmicwhile\ #1}

\DeclareMathSymbol{\mlq}{\mathord}{operators}{``}
\DeclareMathSymbol{\mrq}{\mathord}{operators}{`'}

\begin{document}

\begin{frontmatter}

\title{Efficient Decision Trees for 
Multi-class Support Vector Machines Using Entropy and Generalization Error Estimation}

\author[cu]{Pittipol Kantavat}
\cortext[mycorrespondingauthor]{Corresponding author}
\ead{pittipol.k@student.chula.ac.th}

\author[cu]{Boonserm Kijsirikul\corref{mycorrespondingauthor}}
\ead{boonserm.k@chula.ac.th}
\author[cu]{Patoomsiri Songsiri}
\ead{patoomsiri.s@chula.ac.th}
\author[ou]{Ken-ichi Fukui}
\ead{fukui@ai.sanken.osaka-u.ac.jp}
\author[ou]{Masayuki Numao}
\ead{numao@ai.sanken.osaka-u.ac.jp}
 
\address[cu]{Department of Computer Engineering, Faculty of Engineering, Chulalongkorn University, Thailand}
\address[ou]{Department of Architecture for Intelligence, Division of Information and Quantum Sciences, The Institute of Science and Industrial Research (ISIR), Osaka University, Japan}

\begin{abstract}
We propose new methods for Support Vector Machines (SVMs) using tree architecture for multi-class classification. In each node of the tree, we select an appropriate binary classifier using entropy and generalization error estimation, then group the examples into positive and negative classes based on the selected classifier and train a new classifier for use in the classification phase. The proposed methods can work in time complexity between O(log\textsubscript{2}\textit{N}) to O(\textit{N}) where \textit{N} is the number of classes. We compared the performance of our proposed methods to the traditional techniques on the UCI machine learning repository using 10-fold cross-validation. The experimental results show that our proposed methods are very useful for the problems that need fast classification time or problems with a large number of classes as the proposed methods run much faster than the traditional techniques but still provide comparable accuracy.
\end{abstract}


\end{frontmatter}

\section{Introduction}\label{introduction}

The Support Vector Machine (SVM) \cite{Vapnik1998, Vapnik1999} was originally designed to solve binary classification problems by constructing a hyperplane to separate the two-class data with maximum margin. For dealing with multi-class classification problems, there have been two main approaches. The first approach is to solve a single optimization problem \cite{Vapnik1998, Bredensteiner1999, Crammer2002}, and the second approach is to combine several binary classifiers. Hsu and Lin \cite{Hsu2002} suggested that the second approach may be more suitable for practical use. 

The general techniques of combining several classifiers are the One-Versus-One method (OVO) \cite{Hastie1998, Knerr1990} and the One-Versus-All method (OVA). OVO requires \textit{N}$\times$(\textit{N}--1)/2 binary classifiers for an \textit{N}-class problem. Usually,  OVO determines the target class using the strategy called Max-Wins \cite{Friedman1996} that selects the class with the highest vote of the binary classifiers. OVA requires \textit{N} classifiers for an \textit{N}-class problem. In the OVA training process, the $i^{th}$ classifier employs the $i^{th}$ class data as positive examples and the remaining classes as negative examples. The output classes are determined by selecting the class with the highest classification score. Both techniques are widely used for the classification problems. In this paper, we will base our techniques on the OVO method.

Although OVO generally yields high accuracy results, it takes O($N$\textsuperscript{2}) classification time, and thus it is not suitable to a problem with a large number of classes. To reduce the running time, Decision Directed Acyclic Graph (DDAG) \cite{Platt2000} and Adaptive Directed Acyclic Graph (ADAG) \cite{kijsirikul2002} have been proposed. Both methods build \textit{N}$\times$(\textit{N}--1)/2 binary classifiers but they employ only \textit{N}--1 classifiers to determine the output class. Though both methods reduce the running time to O($N$), their accuracy is usually lower than the accuracy of OVO.

Some techniques apply the decision-tree structure to eliminate more than one candidate class at each node (classifier) of the tree. Fei and Liu have proposed the Binary Tree of SVM \cite{Fei2006} that selects tree node classifiers randomly or selects by the training data centroids. Songsiri et al. have proposed the Information-Based Dichotomization Tree \cite{songsiri2008} that uses entropy for classifier selection. Chen et al. employ the Adaptive Binary Tree \cite{chen2009} that applies the minimization of the average number of support vectors for tree construction. Bala and Agrawal have proposed the Optimal Decision Tree Based Multi-class SVM \cite{Bala2011} that calculates statistical measurement for decision-tree construction. These techniques share a common disadvantage that a selected classifier may not perfectly separate examples of a class to only the positive or negative side, and hence some techniques allow data of a class to be duplicated to more than one node of the tree. There are also some techniques that construct a decision-tree SVM using the data centroid. Takahashi and Abe have proposed a Decision-Tree-Based SVM \cite{Takahashi2002} that calculates the Euclidian distance and Mahalanobis distance as a separability measure for class grouping. Madzarov et al. have proposed SVM Binary Decision Tree \cite{Madzarov2009} using class centroids in the kernel space.

In this paper, we propose two novel techniques for the problem with a large number of classes. The first technique is the {\it Information-Based Decision Tree SVM} that employs entropy to evaluate the quality of OVO classifiers in the node construction process. The second technique is the {\it Information-Based and Generalization-Error Estimation Decision Tree SVM} that enhances the first technique by integrating generalization error estimation. The key mechanism of both techniques is the method called {\it Class-grouping-by-majority}: when a classifier of a tree node cannot perfectly classify examples of any class into either only positive or negative side of the classifier, the method will group the whole examples of that class into only one side that contains the majority of the examples, and then train a new classifier for the node.

We ran experiments comparing our proposed techniques to the traditional techniques using twenty datasets from the UCI machine learning repository, and conducted the significant test using the Wilcoxon Signed Rank Test \cite{Demsar2006}. The results indicate that our proposed methods are useful, especially for problems that need fast classification or problems with a large number of classes.

This paper is organized as follows. Section~\ref{relatedwork} discusses the tree-structure multi-class SVM techniques. Section~\ref{Proposed_Method} proposes our techniques. Section~\ref{ExpAndResults} provides the experimental details. Section~\ref{conclusion} summarizes our work. 

\section{The OVO-based decision tree SVM}\label{relatedwork}

\subsection{Binary Tree of SVM}

The Binary Tree SVM (BTS) \cite{Fei2006} was proposed by Fei and Liu. BTS randomly selects binary classifiers to be used as decision nodes of the tree. As mentioned previously, BTS allows duplicated classes to be scattered in the tree. Basically, data of a class will be duplicated into the left and right child nodes of a decision node when a classifier of the decision node does not completely classify the whole data of the class into only one side (either positive or negative side). An alternative version of BTS, c-BTS excludes the randomness by using data centroids. In the first step, the centroid of all data is calculated. Then, the centroid of each data class and its Euclidean distance to the centroid of all-data are calculated.  Finally, the ($i$~vs~$j$) classifier is selected such that the centroid of class $i$ and the centroid of class $j$ have the nearest distances to the all-data centroid. 

\begin{figure}
       \centering        
        \begin{subfigure}[b]{0.45\textwidth}				
                \includegraphics[width=\textwidth]{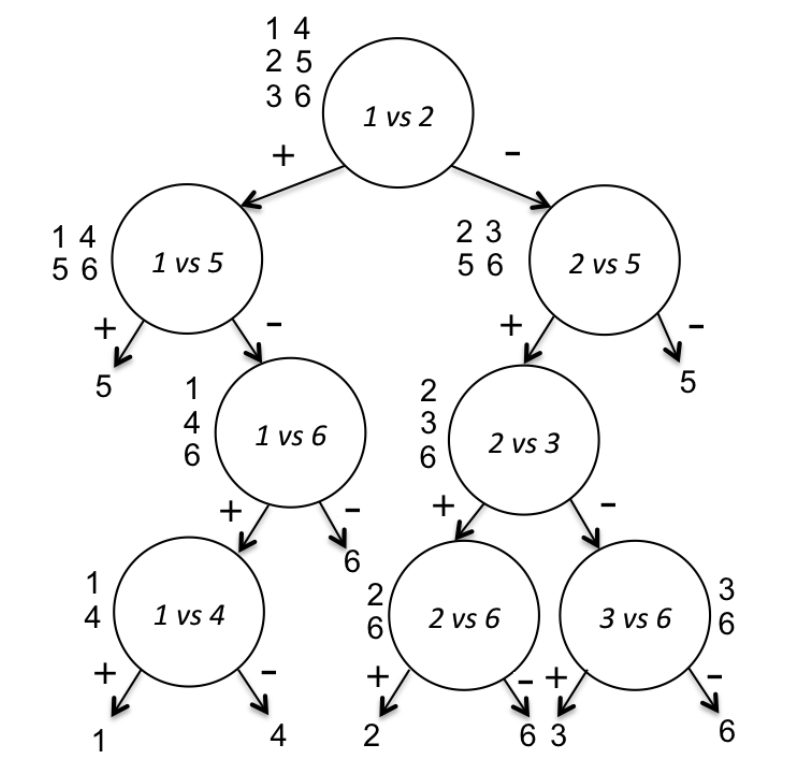}                
                 \hspace{-10pt}
        \end{subfigure}%
 \caption{
\textcolor{black}{The Illustration of the Binary Tree of SVM (BTS). Some data classes may be scattered to several leaf nodes. \cite{Fei2006}}
}
\label{BTS}
\hspace{-100pt}
\end{figure}

The illustrations of BTS and c-BTS are shown in Figure~\ref{BTS}. At the root node, classifier 1~vs~2 is selected. Classes 1, 4 and classes 2, 3 are separated to positive and negative sides, respectively. However, classes 5 and 6 cannot be completely separated. They are reassigned to both positive and negative child nodes. The recursive process continues until finished. Eventually, the duplicated leaf-nodes of classes 5 and 6 appear more than once in the tree.

The classification accuracy and time complexity of BTS and c-BTS may vary according to the threshold configuration. A higher threshold will increase the accuracy but will also increase the running time. The time complexity can be O(log\textsubscript{2}\textit{N}) in the best situation. However, the average time complexity was proven to be log\textsubscript{4/3}((\textit{N}+3)/4)\cite{Fei2006}. 

\subsection{Information-Based Dichotomization}

The Information-Based Dichotomization (IBD) \cite{songsiri2008}, proposed by Songsiri et al., employs information theory to construct a multi-class classification tree. In each node, IBD selects the OVO classifier with minimum entropy. In this method, a data class with a high probability of occurrence will be separated first, and hence this kind of class node will be found in very few levels from the root node. 

IBD also faces the problem that a selected classifier may not perfectly classify the examples with the same class label into only positive or negative side. In this situation, the examples under consideration will be scattered to both positive and negative nodes. To relax this situation, IBD proposes the tree pruning algorithm that ignores the minority examples on the other side of the hyperplane if the percentage of the minority is below a threshold. Applying the tree pruning algorithm to tree construction will eliminate unnecessary duplicated classes in the tree and will decrease the tree depth leading to faster classification speed. However, the tree pruning  algorithm  may risk losing some useful information to the information loss and decrease the classification accuracy.

\section{The proposed methods}\label{Proposed_Method}

We propose two novel techniques that are aimed at achieving high classification speed and may sacrifice classification accuracy to some extent. We expect them to work in the time complexity of O(log\textsubscript{2}\textit{N}) in the best case, and thus the proposed techniques are suitable for the problem with a large number of classes that cannot be solved efficiently in practice by the methods with O($N$\textsuperscript{2}) classification time.

\subsection{The Information-Based Decision Tree}

The Information-Based Decision Tree (IB-DTree) is an OVO-based multi-class classification technique. IB-DTree builds a tree by adding decision nodes one by one; it selects the binary classifier with minimum entropy as the {\it initial classifier} of the decision node. Minimum entropy classifiers will lead to a fast classification time for the decision tree because data classes with high probability of occurrence will be found within a few steps from the root node. The initial classifier will be adjusted further to be the {\it final classifier} for the decision node as described later. The entropy of a binary classifier $h$ can be calculated by Equation~\ref{eq:2}. \textit{p}\textsuperscript{+} and \textit{p}\textsuperscript{--} are the proportions of the positive and negative examples (corresponding to the classifier) to all training examples, respectively. Similarly, $p$\textsubscript{$i$}\textsuperscript{+} is the proportion of positive examples of the class $i$ to all positive examples. $p$\textsubscript{$i$}\textsuperscript{--} is the proportion of negative examples of the class $i$ to all negative examples. In case there is no positive (or negative) example of classifier $h$ for any class, the term ($-p\textsubscript{$i$}\textsuperscript{+} log\textsubscript{2}p\textsubscript{$i$}\textsuperscript{+}$) or ($-p\textsubscript{$i$}\textsuperscript{--} log\textsubscript{2}p\textsubscript{$i$}\textsuperscript{--}$) of that class will be defined as 0. 

\begin{equation} \label{eq:2}
\begin{split}
Entropy(h) = p\textsuperscript{+}\times[\sum_{i=1}^{N} -p\textsubscript{$i$}\textsuperscript{+} log\textsubscript{2}p\textsubscript{$i$}\textsuperscript{+}]
\\ + p\textsuperscript{--}\times[\sum_{i=1}^{N} -p\textsubscript{$i$}\textsuperscript{--} log\textsubscript{2} p\textsubscript{$i$}\textsuperscript{--}]
\end{split}
\end{equation}

From \textit{N}$\times$(\textit{N}--1)/2 OVO-classifiers, the classifier with minimum entropy is selected by using Equation~\ref{eq:2} as the initial classifier $h$. Examples of a specific class may scatter on both positive and negative sides of the initial classifier. The key mechanism of  IB-DTree is the Class-grouping-by-majority method, in Algorithm~\ref{Class-grouping-by-majority-algo}, that groups examples of each class having scattering examples to be on the same side containing the majority of the examples. Using the groups of examples labeled by Class-grouping-by-majority, IB-DTree then trains the final classifier $h'$ of the decision node. A traditional OVO-based algorithm might face the problem when encountering data of another class $k$ using classifier $h$ = ($i$~vs~$j$) and having to scatter examples of class $k$ to both left and right child nodes. Our proposed method will never face this situation, because data of class $k$ will always be grouped in either a positive or negative group of classifier $h'$ = ($P$~vs~$N$). Hence, there is no need to duplicate class $k$ data to the left or right child node, and the tree depth will not be increased unnecessarily.

\begin{algorithm*}[t]
  \caption{Class-grouping-by-majority }\label{Class-grouping-by-majority-algo}
  \begin{algorithmic}[1]
  \Procedure{Class-grouping-by-majority (selected classifier $h$, candidate classes $K$)}{}
  \State Initialize set of positive classes $P$ = $\varnothing$ and set of negative classes $N$ = $\varnothing$ 
  \For{each class $i$ $\in$ $K$}:-
  	\State Label all data of class $i$ to (+) and (--) separated by initial classifier $h$
	\State $p$ $\leftarrow$ count(+), $n$ $\leftarrow$ count(--)
  	\State if ($p$ > $n$) then $P$ $\leftarrow$ $P$ $\cup$ \{$i$\}
	\State else $N$ $\leftarrow$ $N$ $\cup$ \{$i$\}
  \EndFor
  \State Train final classifier $h'$ using all data if classes in $P$ as positive examples, and in $N$ as negative examples
  \State \textbf{return} \textit{h'}, $P$, $N$
  \EndProcedure
  \end{algorithmic}
\end{algorithm*}

\begin{figure}
       \centering        
        \begin{subfigure}[b]{0.45\textwidth}				
                \includegraphics[width=\textwidth]{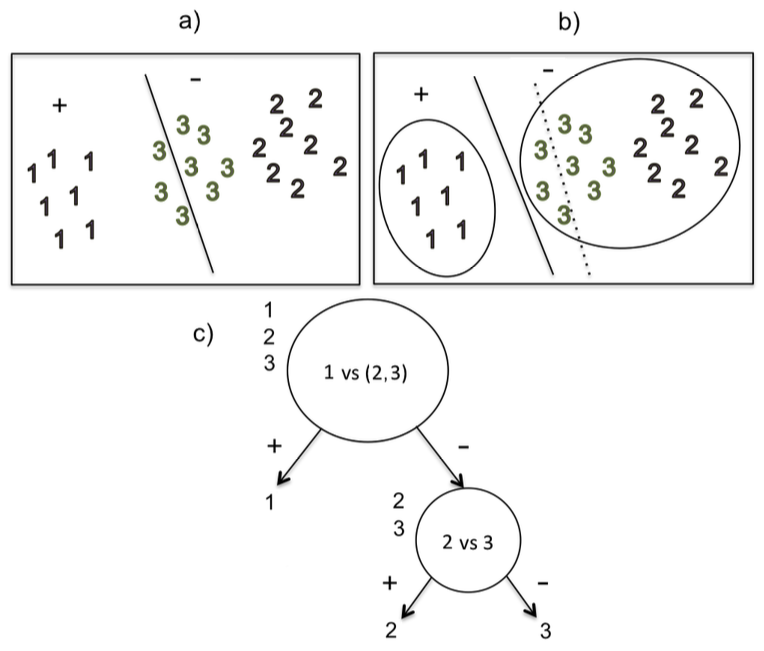}                
                 \hspace{-10pt}
        \end{subfigure}%
 \caption{
\textcolor{black}{An example of the Class-grouping-by-majority strategy. a) before grouping b) after grouping c) the decision tree after grouping.}
}
\label{Class-grouping-by-majority}
\hspace{-100pt}
\end{figure}

An example of the Class-grouping-by-majority for a 3-class problem is shown in Figure~\ref{Class-grouping-by-majority}. Suppose that we select initial classifier 1~vs~2 as \textit{h} for the root node. In Figure~\ref{Class-grouping-by-majority}(a), most of class-3 data is on the negative side of the hyperplane. Therefore, we assign all training data of class-3 as negative examples and train classifier 1~vs~(2, 3) as a new classifier $h'$ for use in the decision tree as in Figure~\ref{Class-grouping-by-majority}(b). As the result, we obtain a decision tree constructed by IB-DTree as shown in Figure~\ref{Class-grouping-by-majority}(c).

\begin{figure}
       \centering        
        \begin{subfigure}[b]{0.45\textwidth}				
                \includegraphics[width=\textwidth]{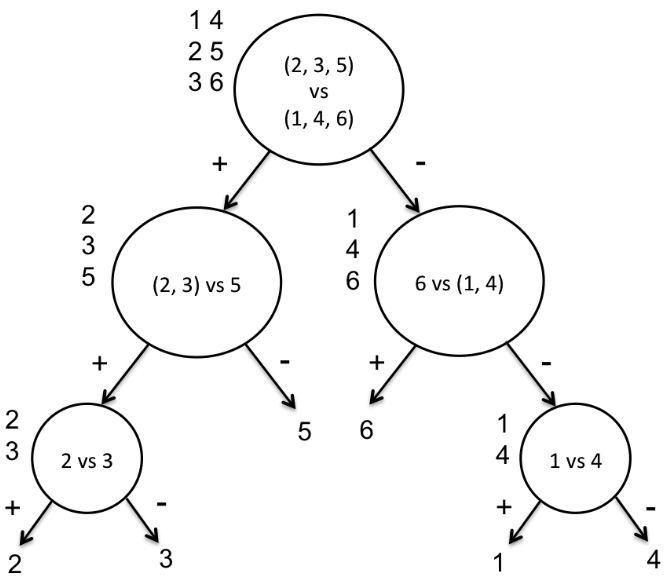}                
                 \hspace{-10pt}
        \end{subfigure}%
 \caption{
\textcolor{black}{The Illustration of the Information-Based Decision Tree (IB-DTree). No duplicated class at the leaf nodes.}
}
\label{IBDTree}
\hspace{-100pt}
\end{figure}

To illustrate more about IB-DTree, we show in Figure~\ref{IBDTree} a decision tree constructed by IB-DTree using the same example as in Figure~\ref{BTS} of the BTS method. At the root node, classifier 1~vs~2 is selected as \textit{h}. Most of the training examples of classes 3 and 5 are on the positive side of \textit{h}, while the majority of training examples of classes 4 and 6 are on the negative side of \textit{h}. Consequently, (2, 3, 5)~vs~(1, 4, 6) is trained as classifier $h'$. For the remaining steps of the tree, the process continues recursively until finished, and there is no duplicated class leaf-node in the tree.

As described in Algorithm~\ref{IB-DTree-algo}, IB-DTree constructs a tree using a recursive procedure starting from the root node from lines 1-7 with all candidate classes in line 2. The node-adding procedure will be processed from lines 8-19. First, the initial classifier $h$ with the lowest entropy will be selected. Second, data of each class will be grouped to either the positive group ($P$) or the negative group ($N$). Then, the final classifier $h'$ will be trained using $P$ and $N$ and will be assigned as decision node. Finally, the algorithm processes the child nodes recursively and stops the process at the leaf nodes when the stopping conditions holds.  

\begin{algorithm*}[t]
\caption{Information-Based Decision Tree SVM (IB-DTree)}\label{IB-DTree-algo}
\begin{algorithmic}[1]
\Procedure{IB-DTree}{}
	\State Initialize the tree $T$ with root node $Root$ 
	\State Initialize the set of candidate output classes $S=\{1, 2, 3,..., N\}$
	\State Create the all binary classifiers ($i$~vs~$j$); $i$, $j$ $\in$ $S$
	\State Construct Tree ($Root$, $S$)
	\State \textbf{return} $T$
\EndProcedure		
\Procedure{Construct Tree (node $D$, candidate classes $K$)}{}
\For{each binary classifier ($i$~vs~$j$); $i$, $j$ $\in$ $K$; $i$ < $j$}:-
  	\State Calculate the entropy using training data of all classes in $K$
\EndFor
\State initial classifier $h$ $\leftarrow$ classifier ($i$~vs~$j$) with the lowest entropy
\State final classifier $h'$, positive classes $P$, negative classes $N$ $\leftarrow$ Class-grouping-by-majority($h$, $K$)
\State $D$.classifier $\leftarrow$ $h'$
\State Initialize new node $L$; $D$.left-child-node $\leftarrow$ $L$
\State Initialize new node $R$; $D$.right-child-node $\leftarrow$ $R$
\State \textbf{if} |$P$| > 1 \textbf{then} Construct Tree ($L$, $P$) \textbf{else} $L$ is the leaf node with answer class $P$
\State \textbf{if} |$N$| > 1 \textbf{then} Construct Tree ($R$, $N$) \textbf{else} $R$ is the leaf node with answer class $N$
\EndProcedure
\end{algorithmic}
\end{algorithm*}

There are several benefits of IB-DTree. First, data class with high probability of occurrence will be found in only few levels from the root node. Second, there is no duplicated class leaf-node and the depth of the tree is small compared to other methods. Finally, there is no information loss because there is no data pruning.

\subsection{The Information-Based and Generalization-Error Estimation Decision Tree}

The Information-Based and Generalization-Error Estimation Decision Tree (IBGE-DTree) is an enhanced version of IB-DTree. 
In the node-building process, IBGE-DTree selects classifiers using both entropy and generalization error estimation. 

The details of IBGE-DTree are described in Algorithm~\ref{IBGE-DTree-algo}. The IBGE-DTree algorithm is different from IB-DTree at lines 12-17. Instead of selecting classifiers based only on lowest entropy, it also considers generalization error of the classifiers. First, IBGE-DTree ranks the classifiers in ascending order by the entropy. Then, it trains some classifiers using the Class-grouping-by-majority technique and selects the classifier with the lowest generalization error. The positive group ($P$) and negative group ($N$) for building the child nodes in lines 22-23 are obtained from the classifier with the lowest generalization error in lines 18-19.

\begin{algorithm*}[t]
\caption{Information-Based and Generalization-Error Estimation Decision Tree SVM (IBGE-DTree)}\label{IBGE-DTree-algo}
\begin{algorithmic}[1]
\Procedure{IBGE-DTree}{}
	\State Initialize the tree $T$ with root node $Root$ 
	\State Initialize the set of candidate output classes $S=\{1, 2, 3,..., N\}$
	\State Create the all binary classifiers ($i$~vs~$j$); $i$, $j$ $\in$ $S$
	\State Construct Tree ($Root$, $S$)
	\State \textbf{return} $T$
\EndProcedure		
\Procedure{Construct Tree (node $D$, candidate classes $K$)}{}
\For{each binary classifiers ($i$~vs~$j$); $i$, $j$ $\in$ $K$; $i$ < $j$}:-
  	\State Calculate the entropy using training data of all classes in $K$
\EndFor
\State Sort the list of the initial classifiers ($i$~vs~$j$) in ascending order by the entropy as $h\textsubscript{1}$... $h\textsubscript{all}$
\For{each initial classifiers $h\textsubscript{k}$; $k=\{1, 2, 3,..., n\}$, $n$ = number of considering classifiers}
	\State final classifier $h\textsubscript{k}'$, positive classes $P\textsubscript{k}$, negative classes $N\textsubscript{k}$  $\leftarrow$ Class-grouping-by-majority($h\textsubscript{k}$, $K$)
	\State calculate generalization error estimation of final classifier $h\textsubscript{k}'$
\EndFor
\State $D$.classifier $\leftarrow$ final classifiers with the lowest generalization error among $h\textsubscript{1}'$... $h\textsubscript{n}'$
\State $P'$ $\leftarrow$ $P$ used for training the final classifier with the lowest generalization error estimation
\State $N'$ $\leftarrow$ $N$ used for training the final classifier with the lowest generalization error estimation
\State Initialize new node $L$; $D$.left-child-node $\leftarrow$ $L$
\State Initialize new node $R$; $D$.right-child-node $\leftarrow$ $R$
\State \textbf{if} |$P'$| > 1 \textbf{then} Construct Tree ($L$, $P'$) \textbf{else} $L$ is the leaf node with answer class $P'$
\State \textbf{if} |$N'$| > 1 \textbf{then} Construct Tree ($R$, $N'$) \textbf{else} $R$ is the leaf node with answer class $N'$

\EndProcedure
\end{algorithmic}
\end{algorithm*}

The generalization error estimation is the evaluation of a learning model actual performance on unseen data. For SVMs, a model is trained using the concept of the structure risk minimization principle \cite{Vapnik1974}. The performance of an SVM is based on the VC dimension of the model and the quality of fitting training data (or empirical error). The expected risk \textit{R}($\alpha$) is bounded by the following equation \cite{Barlett1999, Burges1998} :

\begin{equation} \label{eq:3}
R(\alpha) \le \frac{l}{m} + \sqrt{\frac{c}{m}(\frac{R\textsuperscript{2}}{\Delta\textsuperscript{2}}   log\textsuperscript{2}m+log\frac{1}{\delta})}
\end{equation}

\noindent
where \textit{l}, \textit{R}, $\Delta$ are the number of labeled examples with margin less than $\Delta$, the radius of the smallest sphere that contains all data points, and the distance between the hyperplane and the closest points of the training set (margin size) respectively. The first and second terms of Inequation~\ref{eq:3} define the empirical error and the VC dimension, respectively. 

Generalization error can be estimated directly using k-fold cross-validation and used to compare the performance of binary classifiers, but it consumes a high computational cost. Another method to estimate the generalization error is by using Inequation \ref{eq:3} with the appropriate parameter substitution \cite{Songsiri2015}. Using the latter method, we can compare relative generalization error on the same datasets and environments. In Section~\ref{ExpAndResults}, we set the value of $c$ = 0.1 and $\delta$ = 0.01 in the experiments. 

As IBGE-DTree is an enhanced version of IB-DTree, its benefits are very similar to the benefits of IB-DTree. However, as it combines the generalization error estimation with the entropy, the selected classifiers are more effective than IB-DTree. 

\section{Experiments and Results}\label{ExpAndResults}

We performed the experiments to compare our proposed methods, IB-DTree and IBGE-DTree, to the traditional strategies, i.e., OVO, OVA, DDAG, ADAG, BTS-G and c-BTS-G. 

We ran experiments based on 10-fold cross-validation on twenty datasets from the UCI repository \cite{Blake1998}, as shown in Table~\ref{dataset}. For the datasets containing both training data and test data, we merged the data into a single set, and then we used 10-fold cross validation to evaluate the classification accuracy. We normalized the data to the range [-1, 1]. We used the software package SVMlight version 6.02 \cite{Joachims2008}. The binary classifiers were trained using the RBF kernel. The suitable kernel parameter ($\gamma$) and regularization parameter \textit{C} for each dataset were selected from \{0.001, 0.01, 0.1, 1, 10\} and \{1, 10, 100, 1000\}, respectively.

To compare the performance of IB-DTree and IBGE-DTree to the other tree-structure techniques, we also implemented BTS-G and c-BTS-G that are our enhanced versions of BTS and c-BTS \cite{Fei2006} by applying Class-grouping-by-majority to improve efficiency of the original BTS and c-BTS. For BTS-G, we selected the classifier for each node randomly 10 times and calculated the average results. For c-BTS-G, we selected the pairwise classifiers in the same way to the original c-BTS.

For DDAG and ADAG where the initial order of classes can affect the final classification accuracy, we examined all datasets by randomly selecting 50,000 initial orders and calculated the average classification accuracy. For IBGE-DTree, we set $n$ (the number of considering classifiers in line 13 of Algorithm~\ref{IBGE-DTree-algo}) to 20 percent of all possible classifiers. For example, if there are 10 classes to be determined, the number of all possible classifiers will be 45. Thus, the value of \textit{n} will be 9.

\begin{table}[]
\centering
\caption{The experimental dataset.}
\small\addtolength{\tabcolsep}{-4pt}
{\scriptsize
\label{dataset}
\begin{tabular}{lccr}
\hline
Dataset Name &   \#Classes &   \#Attributes &  \#Examples \\
\hline
Page Block  &   5 &   10 &  5473\\
Segment &   7 &   18 &  2310\\
Shuttle &   7 &   9 &  58000\\
Arrhyth &   9 &   255 &  438\\
Cardiotocography &   10 &   21 &  2126\\
Mfeat-factor &   10 &   216 &  2000\\
Mfeat-fourier &   10 &   76 &  2000\\
Mfeat-karhunen &   10   & 64 &  2000\\
Optdigit &   10 &   62 &  5620\\
Pendigit &   10 &   16 &  10992\\
Primary Tumor &   13 &   15 &  315\\
Libras Movement &   15 &   90 &  360\\
Abalone &   16 &   8 &  4098\\
Krkopt &   18 &   6 &  28056\\
Spectrometer &   21 &   101 &  475\\
Isolet &   26 &   34 &  7797\\
Letter &   26 &   16 &  20052\\
Plant Margin &   100 &   64 &  1600\\
Plant Shape &   100 &   64 &  1600\\
Plant Texture &   100 &   64 &  1599\\
\hline
\end{tabular}
}
\end{table}

The experimental results are shown in Tables~\ref{accuracy_results}-~\ref{decision_time}. Table~\ref{accuracy_results} presents the average classification accuracy results of all datasets, and Table~\ref{significant_test} shows the Wilcoxon Signed Rank Test \cite{Demsar2006} to assess the accuracy of our methods with others. Table~\ref{decision_time} shows the average decision times that are used to determine the output class of a test example.

In Table~\ref{accuracy_results}, a bold number indicates the highest accuracy in each dataset. The number in the parentheses shows the ranking of each technique. The highest accuracy is obtained by OVO, followed by ADAG, OVA and DDAG. Among the tree structure techniques, IBGE-DTree yields the highest accuracy, followed by IB-DTree, BTS-G and c-BTS-G. 

Table~\ref{significant_test} shows a significant difference between techniques in Table~\ref{accuracy_results} using the Wilcoxon Signed Rank Test. The bold numbers indicate the significant win (or loss) with the significance level of 0.05. The numbers in the parentheses indicate the pairwise win-lose-draw between the techniques under comparison. The statistical tests indicate that OVO significantly outperforms all other techniques. Among the tree structure techniques, IBGE-Dtree provides the highest accuracy results. IBGE-DTree is also insignificantly different from OVA, DDAG, ADAG and IB-DTree. BTS-G and c-BTS-G significantly underperform the other techniques. 
	
Table~\ref{decision_time} shows the average number of decisions required to determine the output class of a test example. The lower the average number of decisions, the faster the classification speed. IB-DTree and IBGE-DTree are the fastest among the techniques compared, while OVO is the slowest one.

\begin{table*}[htp]
\centering
\caption[]{The average classification accuracy results and their standard deviation. A bold number indicates the highest accuracy in each dataset. The numbers in the parentheses show the accuracy ranking.}
\small\addtolength{\tabcolsep}{-4pt}
{\scriptsize

\begin{tabular}{ ll ccl ccl ccl ccl }
\hline
Datasets   && \multicolumn{2}{c}{OVA}  && \multicolumn{2}{c}{OVO}  && \multicolumn{2}{c}{DDAG} && \multicolumn{2}{c}{ADAG} & \\
\hline
Page Block    && \textbf{96.857 $\pm$ 0.478} & \textbf{(1)} &
		& 96.735 $\pm$ 0.760 & (3) &
		& 96.729 $\pm$ 0.764 & (4) &
		& 96.740 $\pm$ 0.757 & (2) &\\
Segment    && 97.359 $\pm$ 1.180 & (5) &
		& 97.431 $\pm$ 0.860 & (3) &
		& \textbf{97.442 $\pm$ 0.848} & \textbf{(1)} &
		& 97.436 $\pm$ 0.854 & (2) & \\
Shuttle    	&& 99.914 $\pm$ 0.053 & (5) &
		& \textbf{99.920 $\pm$ 0.054} & \textbf{(1)} &
		& \textbf{99.920 $\pm$ 0.054} & \textbf{(1)} &
		& \textbf{99.920 $\pm$ 0.054} & \textbf{(1)} & \\
Arrhyth    	&& 72.603 $\pm$ 7.041 & (2) &
		& \textbf{73.146 $\pm$ 6.222} & \textbf{(1)} &
		& 67.375 $\pm$ 7.225 & (8) &
		& 67.484 $\pm$ 7.318 & (7) & \\
Cardiotocography    	&& 83.208 $\pm$ 1.661 & (5) &
		& \textbf{84.431 $\pm$ 1.539} & \textbf{(1)} &
		& 84.241 $\pm$ 1.609 & (3) &
		& 84.351 $\pm$ 1.607 & (2) & \\
Mfeat-Factor    && \textbf{98.200 $\pm$ 1.033} & \textbf{(1)} &
		& 98.033 $\pm$ 0.908 & (3) &
		& 98.011 $\pm$ 0.941 & (5) &
		& 98.019 $\pm$ 0.919 & (4) & \\
Mfeat-fourier    && 84.850 $\pm$ 1.528 & (6) &
		& \textbf{85.717 $\pm$ 1.603} & \textbf{(1)} &
		& 85.702 $\pm$ 1.589 & (3) &
		& 85.708 $\pm$ 1.585 & (2) & \\
Mfeat-Karhunen    	&& \textbf{98.000 $\pm$ 0.943} & \textbf{(1)} &
		& 97.913 $\pm$ 0.750 & (3) &
		& 97.894 $\pm$ 0.726 & (5) &
		& 97.900 $\pm$ 0.722 & (4) & \\
Optdigit    	&& 99.324 $\pm$ 0.373 & (2) &
		& \textbf{99.964 $\pm$ 0.113} & \textbf{(1)} &
		& 99.288 $\pm$ 0.346 & (3) &
		& 99.288 $\pm$ 0.346 & (3) & \\
Pendigit    && 99.554 $\pm$ 0.225 & (4) &
		& \textbf{99.591 $\pm$ 0.203} & \textbf{(1)} &
		& 99.569 $\pm$ 0.213 & (3) &
		& 99.574 $\pm$ 0.211 & (2) & \\
Primary Tumor   && 46.667 $\pm$ 7.011 & (3)&
		& \textbf{50.212 $\pm$ 7.376} & \textbf{(1)} &
		& 39.278 $\pm$ 6.419 & (8) &
		& 39.486 $\pm$ 6.483 & (7) & \\
Libras Movement   	&& \textbf{90.000 $\pm$ 2.986} & \textbf{(1)} &
		& 89.074 $\pm$ 3.800 & (2) &
		& 89.034 $\pm$ 3.729 & (3) &
		& 89.017 $\pm$ 3.687 & (4) & \\
Abalone	&& 16.959 $\pm$ 2.388 & (8) &
		& \textbf{28.321 $\pm$ 1.516} & \textbf{(1)} &
		& 24.093 $\pm$ 3.044 & (7) &
		& 24.258 $\pm$ 3.154 & (6) & \\
Krkopt	&& \textbf{85.750 $\pm$ 0.769} & \textbf{(1)} &
		& 82.444 $\pm$ 0.628 & (2) &
		& 81.952 $\pm$ 0.643 & (4) &
		& 82.235 $\pm$ 0.634 & (3) & \\
Spectrometer 	&& 51.579 $\pm$ 6.256 & (8) &
		& \textbf{68.421 $\pm$ 5.007} & \textbf{(1)} &
		& 68.052 $\pm$ 4.706 & (4) &
		& 68.392 $\pm$ 4.796 & (2) & \\
Isolet	&& \textbf{94.947 $\pm$ 0.479} & \textbf{(1)} &
		& 94.898 $\pm$ 0.648 & (2) &
		& 94.872 $\pm$ 0.631 & (4) &
		& 94.885 $\pm$ 0.643 & (3) & \\
Letter 	&& 97.467 $\pm$ 0.305 & (4) &
		& \textbf{97.813 $\pm$ 0.382} & \textbf{(1)} &
		& 97.746 $\pm$ 0.357 & (3) &
		& 97.787 $\pm$ 0.360 & (2) & \\
Plant Margin    && 82.875 $\pm$ 2.655 & (4) &
		& \textbf{84.401 $\pm$ 2.426} & \textbf{(1)} &
		& 84.238 $\pm$ 2.516 & (3) &
		& 84.341 $\pm$ 2.607 & (2) & \\
Plant Shape    && 70.938 $\pm$ 2.783 & (3) &
		& \textbf{71.182 $\pm$ 3.295} & \textbf{(1)} &
		& 70.922 $\pm$ 3.393 & (4) &
		& 71.090 $\pm$ 3.313 & (2) & \\
Plant Texture    && \textbf{87.179 $\pm$ 2.808} & \textbf{(1)} &
		& 86.387 $\pm$ 2.374 & (2) &
		& 86.173 $\pm$ 2.519 & (4) &
		& 86.259 $\pm$ 2.510 & (3) & \\
\hline
Avg. Rank && 3.35 &&
		& 1.70 &&
		& 4.00 &&
		& 3.15 && \\
\hline
\end{tabular}

\begin{tabular}{ ll lll lll ccl ccl }
\hline
Datasets    && \multicolumn{2}{c}{BTS-G}  && \multicolumn{2}{c}{c-BTS-G}  && \multicolumn{2}{c}{IB-DTree} && \multicolumn{2}{c}{IBGE-DTree} &\\
\hline
Page Block    && 96.622 $\pm$ 0.812 & (5) &
		& 96.565 $\pm$ 0.884 & (8) &
		& 96.565 $\pm$ 0.779 & (7) &
		& 96.620 $\pm$ 0.852 & (6) & \\
Segment    && 97.273 $\pm$ 1.076 & (7) &
		& 97.100 $\pm$ 0.957 & (8) &
		& 97.316 $\pm$ 0.838 & (6) &
		& 97.403 $\pm$ 1.100 & (4) & \\
Shuttle    	&& 99.914 $\pm$ 0.050 & (5) &
		& 99.914 $\pm$ 0.053 & (5) &
		& 99.916 $\pm$ 0.050 & (4) &
		& 99.910 $\pm$ 0.053 & (8) & \\
Arrhyth    	&& 71.918 $\pm$ 5.688 & (4) &
		& 71.918 $\pm$ 5.189 & (4) &
		& 71.005 $\pm$ 5.836 & (6) &
		& 72.146 $\pm$ 4.043 & (3) & \\
Cardiotocography    	&& 83.048 $\pm$ 2.147 & (7) &
		& 82.926 $\pm$ 2.106 & (8) &
		& 83.819 $\pm$ 1.710 & (4) &
		& 83.161 $\pm$ 2.490 & (6) & \\
Mfeat-Factor    && 97.810 $\pm$ 0.882 & (8) &
		& 98.000 $\pm$ 0.888 & (6) &
		& 98.000 $\pm$ 0.768 & (6) &
		& \textbf{98.200 $\pm$ 0.816} & \textbf{(1)} & \\
Mfeat-fourier    && 84.235 $\pm$ 1.636 & (8) &
		& 84.350 $\pm$ 1.700 & (7) &
		& 85.200 $\pm$ 1.605 & (4) &
		& 85.150 $\pm$ 1.717 & (5) & \\
Mfeat-Karhunen    	&& 97.450 $\pm$ 0.832 & (6) &
		& 97.050 $\pm$ 0.725 & (8) &
		& 97.450 $\pm$ 0.879 & (6) &
		& 97.950 $\pm$ 1.141 & (2) & \\
Optdigit    	&& 99.002 $\pm$ 0.266 & (8) &
		& 99.039 $\pm$ 0.308 & (7) &
		& 99.164 $\pm$ 0.288 & (5) &
		& 99.093 $\pm$ 0.395 & (6) & \\
Pendigit    && 99.442 $\pm$ 0.184 & (7) &
		& 99.427 $\pm$ 0.201 & (8) &
		& 99.445 $\pm$ 0.198 & (6) &
		& 99.454 $\pm$ 0.318 & (5) & \\
Primary Tumor   && 43.016 $\pm$ 3.824 & (5) &
		& 40.635 $\pm$ 4.813 & (6) &
		& 47.937 $\pm$ 4.567 & (2) &
		& 44.762 $\pm$ 5.478 & (4) & \\
Libras Movement   	&& 87.861 $\pm$ 4.151 & (8) &
		& 88.611 $\pm$ 3.715 & (5) &
		& 88.056 $\pm$ 3.479 & (6) &
		& 88.056 $\pm$ 3.057 & (6) & \\
Abalone	&& 26.635 $\pm$ 1.236 & (3) &
		& 26.013 $\pm$ 1.218 & (4) &
		& 25.281 $\pm$ 0.904 & (5) &
		& 26.745 $\pm$ 0.809 & (2) & \\
Krkopt	&& 77.137 $\pm$ 0.880 & (8) &
		& 78.190 $\pm$ 0.783 & (7) &
		& 79.006 $\pm$ 0.792 & (6) &
		& 80.610 $\pm$ 1.039 & (5) & \\
Spectrometer 	&& 59.432 $\pm$ 5.563 & (6) &
		& 52.421 $\pm$ 5.192 & (7) &
		& 68.211 $\pm$ 3.397 & (3) &
		& 67.789 $\pm$ 6.296 & (5) & \\
Isolet	&& 92.850 $\pm$ 0.799 & (7) &
		& 92.677 $\pm$ 0.702 & (8) &
		& 93.639 $\pm$ 0.261 & (6) &
		& 94.011 $\pm$ 0.640 & (5) & \\
Letter 	&& 96.174 $\pm$ 0.321 & (7) &
		& 96.369 $\pm$ 0.423 & (6) &
		& 96.135 $\pm$ 0.312 & (8) &
		& 96.409 $\pm$ 0.344 & (5) & \\
Plant Margin    && 77.994 $\pm$ 1.946 & (8) &
		& 78.188 $\pm$ 2.739 & (7) &
		& 80.563 $\pm$ 3.638 & (5) &
		& 79.313 $\pm$ 2.863 & (6) & \\
Plant Shape    && 63.219 $\pm$ 1.808 & (7) &
		& 61.750 $\pm$ 3.594 & (8) &
		& 67.000 $\pm$ 2.853 & (5) &
		& 66.750 $\pm$ 2.408 & (6) & \\
Plant Texture    && 78.893 $\pm$ 2.547 & (7) &
		& 78.174 $\pm$ 3.997 & (8) &
		& 80.425 $\pm$ 3.602 & (6) &
		& 80.863 $\pm$ 2.828 & (5) & \\
\hline
Avg. Rank && 6.55 &&
		& 6.85 &&
		& 5.35 &&
		& 4.75 && \\
\hline
\end{tabular}
}

\label{accuracy_results}
\end{table*}

\begin{table*}[htp]
\centering
\caption[]{The significance test of the average classification accuracy results. A bold number means that the result is a significant win (or loss) using the Wilcoxon Signed Rank Test. The numbers in the parentheses indicate the win-lose-draw between the techniques under comparison.}
\small\addtolength{\tabcolsep}{-4pt}
{\scriptsize
\begin{tabular}{ l cc cc cc cc cc cc cc cc}
\hline
		 & OVO  && DDAG && ADAG &
		 & BTS-G  && c-BTS-G  && IB-DTree && IBGE-DTree &\\
\hline
OVA		& 0.1260 &
		& 0.7642 &
		& 1.0000 &
		& \textbf{0.0160} &	
		& \textbf{0.0061} &
		& 0.1443 &
		& 0.0536 & \\
		& (7-13-0) &
		& (11-9-0) &
		& (10-10-0) &
		& \textbf{(17-2-1)} &
		& \textbf{(17-2-1)} &
		& (14-6-0) &
		& (15-4-1) & \\
\hline
OVO		& - &
		& \textbf{0.0002} &
		& \textbf{0.0002} &
		& \textbf{0.0001} &
		& \textbf{0.0001} &
		& \textbf{0.0001} &
		& \textbf{0.0003} & \\
		& - &
		& \textbf{(11-8-1)} &
		& \textbf{(17-2-1)} &
		& \textbf{(20-0-0)} &
		& \textbf{(20-0-0)} &
		& \textbf{(20-0-0)} &
		& \textbf{(18-2-0)} & \\
\hline
DDAG	& - &
		& - &
		& \textbf{0.0014} &
		& \textbf{0.0188} &
		& \textbf{0.0151} &
		& 0.0574 &
		& 0.1010 & \\
		& - &
		& - &
		& \textbf{(2-16-2)} &
		& \textbf{(17-3-0)} &
		& \textbf{(17-3-0)} &
		& (16-4-0) &
		& (15-5-0) & \\
\hline
ADAG	& - &
		& - &
		& - &
		& \textbf{0.0188} &
		& \textbf{0.0124} &
		& \textbf{0.0332} &
		& 0.0536 & \\
		& - &
		& - &
		& - &
		& \textbf{(17-3-0)} &
		& \textbf{(17-3-0)} &
		& \textbf{(17-3-0)} &
		& (15-5-0) & \\
\hline
BTS-G	& - &
		& - &
		& - &
		& - &
		& 0.2846 &
		& \textbf{0.0264} &
		& \textbf{0.0001} & \\
		& - &
		& - &
		& - &
		& - &
		& (12-7-1) &
		& \textbf{(4-15-1)} &
		& \textbf{(2-18-0)} & \\
\hline
c-BTS-G	& - &
		& - &
		& - &
		& - &
		& - &
		& \textbf{0.0466} &
		& \textbf{0.0004} & \\
		& - &
		& - &
		& - &
		& - &
		& - &
		& \textbf{(4-15-1)} &
		& \textbf{(2-18-0)} & \\
\hline
IB-DTree	& - &
		& - &
		& - &
		& - &
		& - &
		& - &
		& 0.4715 & \\
		& - &
		& - &
		& - &
		& - &
		& - &
		& - &
		& (8-11-1)& \\
\hline
\end{tabular}
}
\label{significant_test}
\end{table*}

\begin{table*}[htp]
\centering
\caption[]{The average number of decision times. A bold number indicates the lowest decision times in each dataset.}
\small\addtolength{\tabcolsep}{-4pt}
{\scriptsize
\begin{tabular}{ l cc cc cc cc cc cc cc cc}
\hline
Datasets   & OVA  && OVO  && DDAG && ADAG &
		 & BTS-G  && c-BTS-G  && IB-DTree && IBGE-DTree &\\
\hline
Page Block    & 5 &
		& 10 &
		& 4 &
		& 4 &
		& \textbf{3.628} &
		& 3.801 &
		& 3.790 &
		& 3.831 & \\
Segment    & 7 &
		& 21 &
		& 6 &
		& 6 &
		& 3.630 &
		& 3.882 &
		& \textbf{2.858} &
		& 3.009 & \\
Shuttle    & 7 &
		& 21 &
		& 6 &
		& 6 &
		& \textbf{4.703} &
		& 5.370 &
		& 5.000 &
		& 5.019 & \\
Arrhyth    & 9 &
		& 36 &
		& 8 &
		& 8 &
		& 6.434 &
		& 5.473 &
		& \textbf{5.258} &
		& 5.418 & \\
Cardiotocography    	& 10 &
		& 45 &
		& 9 &
		& 9 &
		& 4.993 &
		& 3.698 &
		& \textbf{3.490} &
		& 3.807 & \\
Mfeat-factor    & 10 &
		& 45 &
		& 9 &
		& 9 &
		& 4.224 &
		& 3.643 &
		& \textbf{3.473} &
		& 3.754 & \\
Mfeat-fourier    & 10 &
		& 45 &
		& 9 &
		& 9 &
		& 4.512 &
		& 3.796 &
		& \textbf{3.522} &
		& 3.786 & \\
Mfeat-karhunen    & 10 &
		& 45 &
		& 9 &
		& 9 &
		& 4.322 &
		& 4.561 &
		& \textbf{3.435} &
		& 3.859 & \\
Optdigit    & 10 &
		& 45 &
		& 9 &
		& 9 &
		& 4.503 &
		& 4.470 &
		& \textbf{3.399} &
		& 4.566 & \\
Pendigit    & 10 &
		& 45 &
		& 9 &
		& 9 &
		& 4.031 &
		& 3.494 &
		& \textbf{3.487} &
		& 3.491 & \\
Primary Tumor    & 13 &
		& 78 &
		& 12 &
		& 12 &
		& 6.672 &
		& 6.476 &
		& \textbf{5.391} &
		& 7.610 & \\
Libras Movement   & 15 &
		& 105 &
		& 14 &
		& 14 &
		& 5.493 &
		& 5.114 &
		& \textbf{4.325} &
		& 4.411 & \\
Abalone    & 16 &
		& 120 &
		& 15 &
		& 15 &
		& 9.242 &
		& 8.540 &
		& 8.768 &
		& \textbf{7.626} & \\
Krkopt    & 18 &
		& 153 &
		& 17 &
		& 17 &
		& 6.743 &
		& 4.847 &
		& \textbf{3.957} &
		& 5.083 & \\
Spectrometer    & 21 &
		& 210 &
		& 20 &
		& 20 &
		& 6.728 &
		& 6.080 &
		& \textbf{4.411} &
		& 4.613 & \\
Isolet    & 26 &
		& 325 &
		& 25 &
		& 25 &
		& 6.865 &
		& 6.015 &
		& \textbf{5.064} &
		& 5.323 & \\
Letter    & 26 &
		& 325 &
		& 25 &
		& 25 &
		& 6.771 &
		& 7.104 &
		& \textbf{4.922} &
		& 5.910 & \\
Plant Margin    & 100 &
		& 4950 &
		& 99 &
		& 99 &
		& 11.338 &
		& 8.600 &
		& \textbf{6.973} &
		& 7.576 & \\
Plant Shape    & 100 &
		& 4950 &
		& 99 &
		& 99 &
		& 11.935 &
		& 9.653 &
		& \textbf{6.965} &
		& 7.446 & \\
Plant Texture    & 100 &
		& 4950 &
		& 99 &
		& 99 &
		& 12.230 &
		& 9.618 &
		& \textbf{7.022} &
		& 8.329 & \\
\hline
\end{tabular}
}
\label{decision_time}
\end{table*}

The experiments show that IBGE-DTree is the most efficient technique among the tree-structure methods. It outputs the answer very fast and provides accuracy comparable to OVA, DDAG and ADAG. IBGE-DTree also performs significantly better than BTS-G and c-BTS-G. OVO yields the highest accuracy among the techniques compared. However, OVO consumes a very high running time for classification, especially when applied to the problems with a large number of classes. For example, for datasets Plant Margin, Plant Shape, and Plant Texture, OVO needs the decision times of 4,950, while IBGE-DTree requires the decision times of only 7.4 to 8.3.

IB-DTree is also a time-efficient technique that yields the lowest average decision times but gives lower classification accuracy than IBGE-Tree. The classification accuracy of IB-DTree is comparable to OVA, DDAG and significantly better than BTS-G and c-BTS-G, but it significantly underperforms OVO and ADAG. Although in the general case IBGE-DTree is more considerable than IB-DTree because it yields better classification accuracy, IB-DTree is an interesting option when the training time is the main concern. 

\section{Conclusions}\label{conclusion}

In this research, we proposed IB-DTree and IBGE-DTree, the techniques that combine the entropy and the generalization error estimation for the classifier selection in the tree construction. Using the entropy, the class with high probability of occurrence will be placed near the root node, resulting in reduction of decision times for that class. The lower the number of decision times, the less the cumulative error of the prediction because every classifier along the path may give a wrong prediction. The generalization error estimation is a method for evaluating the effectiveness of the binary classifier. Using generalization error estimation, only accurate classifiers are considered for use in the decision tree. Class-grouping-by-majority is also a key mechanism to the success of our methods that is used to construct the tree without duplicated class scattering in the tree. Both IB-DTree and IBGE-DTree classify the answer in the decision times of no more than O(\textit{N}).

We performed the experiments comparing our methods to the traditional techniques on twenty datasets from the UCI repository. We can summarize that IBGE-DTree is the most efficient technique that gives the answer very fast; provides accuracy comparable to OVA, DDAG, and ADAG; and yields better accuracy than the other tree-structured techniques. IB-DTree also works fast and provides accuracy comparable to IBGE-DTree and could be considered when training time is the main concern.

\section{Acknowledgments}\label{Acknowledgment}
This research was supported by The Royal Golden Jubilee Ph.D Program and The Thailand Research Fund.

\section{References}\label{refer}

\footnotesize

\bibliographystyle{elsarticle-harv}
\bibliography{IBTreeRef}

\end{document}